\def\papertitle{Cross-modal variational inference for bijective signal-symbol translation}

\def\paperauthorA{Axel Chemla--Romeu-Santos\textsuperscript{1,2}, Stavros Ntalampiras\textsuperscript{1}, Philippe Esling\textsuperscript{2}, Goffredo Haus\textsuperscript{1}, G\'{e}rard Assayag\textsuperscript{1}}

\documentclass[twoside,a4paper]{article}
\usepackage{dafx_19}
\usepackage{amsmath,amssymb,amsfonts,amsthm}
\usepackage{euscript}
\usepackage{inputenc}
\usepackage[T1]{fontenc}
\usepackage{ifpdf}
 \usepackage{balance} 
\usepackage[english]{babel}
\usepackage{caption}
\usepackage{subfig} 
\usepackage{color}
\usepackage{multicol}
\usepackage{mdwtab}

\ninept

\usepackage{times}

\newif\ifpdf
\ifx\pdfoutput\relax
\else
   \ifcase\pdfoutput
      \pdffalse
   \else
      \pdftrue
\fi

\ifpdf 
  \usepackage[pdftex,
    pdftitle={\papertitle},
    pdfauthor={\paperauthorA},
    colorlinks=false, 
    bookmarksnumbered, 
    pdfstartview=XYZ 
  ]{hyperref}
  \usepackage[pdftex]{graphicx}
\else 
  \usepackage[dvips]{epsfig,graphicx}
  \usepackage[dvips,
    colorlinks=false, 
    bookmarksnumbered, 
    pdfstartview=XYZ 
  ]{hyperref}
\fi
  \usepackage[hypcap=true]{caption}
\title{\papertitle}
\affiliation{
\paperauthorA }
{ \begin{multicols}{2} \textsuperscript{1} Laboratorio d'Informatica Musicale (LIM) \\ UNIMI, Milano, Italy \\ { \tt \href{mailto:Axel.Chemla@unimi.it}{axel.chemla@unimi.it}} { \tt \href{mailto:stavros.ntalampiras@unimi.it}{stavros.ntalampiras@unimi.it}} { \tt \href{mailto:goffredo.haus@unimi.it}{goffredo.haus@unimi.it}} \\ \columnbreak \textsuperscript{2} IRCAM - CNRS UMR 9912 \\ Sorbonne Universit\'{e}, Paris, France \\ { \tt \href{mailto:esling@ircam.fr}{esling@ircam.fr}} { \tt \href{mailto:esling@ircam.fr}{assayag@ircam.fr}}  \end{multicols}  }

\begin{document}
\ifpdf 
  \DeclareGraphicsExtensions{.png,.jpg,.pdf}
\else  
  \DeclareGraphicsExtensions{.eps}
\fi

\maketitle

\begin{abstract}
Extraction of symbolic information from signals is an active field of research enabling numerous applications especially in the Musical Information Retrieval domain. This complex task, that is also related to other topics such as pitch extraction or instrument recognition, is a demanding subject that gave birth to numerous approaches, mostly based on advanced signal processing-based algorithms. However, these techniques are often non-generic, allowing the extraction of definite physical properties of the signal (pitch, octave), but not allowing arbitrary vocabularies or more general annotations. On top of that, these techniques are one-sided, meaning that they can extract symbolic data from an audio signal, but cannot perform the reverse process and make symbol-to-signal generation. 
In this paper, we propose an bijective approach for signal/symbol translation by turning this problem into a density estimation task over signal and symbolic domains, considered both as related random variables. We estimate this joint distribution with two different variational auto-encoders, one for each domain, whose inner representations are forced to match with an additive constraint, allowing both models to learn and generate separately while allowing signal-to-symbol and symbol-to-signal inference. In this article, we test our models on pitch, octave and dynamics symbols, which comprise a fundamental step towards music transcription and label-constrained audio generation. In addition to its versatility, this system is rather light during training and generation while allowing several interesting creative uses that we outline at the end of the article.
\end{abstract}

\section{Introduction}\label{sec:introduction}
Music Information Retrieval (MIR) is a growing domain of audio processing that aims to extract information (labels, symbolic or temporal features) from audio signals \cite{downie2003music,8665366}. This field embeds both musical and scientific challenges paving the way to a large variety of tasks. Such abundant industrial and creative applications \cite{casey2008content} have attracted the interest of a large number of researchers with plentiful results. Among the diverse sub-tasks included in MIR, \textit{music transcription} comprises an active research field  \cite{klapuri2007signal, benetos2013automatic} which is not only interesting by itself but finds generic applicability as a sub-task for other MIR objectives (cover recognition, key detection, symbolic analysis). Music transcription can be described as associating symbols to audio signals composed of one or more musical instruments. Thus, this field embeds pitch and multi-pitch estimation tasks but also other musical dimensions, such as dynamics. Currently, most pitch estimation techniques are based on fundamental frequency detection \cite{de2002yin}. However, such approaches may prove insufficient in multi-pitch contexts, where the need for more sophisticated approaches appears crucial. 

In parallel, the recent rise of \textit{generative systems} provided interesting alternatives to supervised machine learning approaches focusing on classification \cite{bengio2013generalized}. These unsupervised learning models aim to discover the inner structure of a dataset based on a reconstruction task. Such methods are usually defined as probabilistic density estimation approaches, Bayesian inference and auto-encoding structures. Among those, the \textit{Variational Auto-Encoders} (VAE) provides a powerful framework, which explicitly targets the construction of a \textit{latent space} \cite{kingma2013auto}. Such spaces are high-level representations with the ability to reveal interesting properties about the inner structure of different types of data \cite{kingma2013auto}\cite{rezende2014stochastic}, and also more recently in audio \cite{esling2018generative}. Such learning procedures can be mixed with supervised learning to perform label extraction and conditional generation, showing the flexibility and the efficiency of this approach. Last but not least, latent spaces can also be explicitly shared by several systems acting on different data domains, providing an elegant way of performing domain-to-domain translation or multi-modal learning \cite{liu2017unsupervised}.\\

In this article, we propose a generative modeling approach to musical transcription by formulating it as a density estimation problem. Our approach allows to directly model pairs ($\mathbf{x, y}$), where $\mathbf{x}$ represents the spectral features and $\mathbf{y}$ represents the corresponding musical annotations. Following a multi-modal approach inspired by Higgins \& al. \cite{higgins2017scan}, we train two different VAEs on these separate domains whose latent representations are progressively shared through explicit distribution matching. In addition to providing a Bayesian formulation of musical transcription compatible with arbitrary vocabularies, our method also naturally handles the reverse audio generation process, and thus allows both \textit{signal-to-symbol} and \textit{symbol-to-signal} inference. Furthermore, direct data/symbol generation is also available by latent space exploration, providing an interesting method for creative audio synthesis. Finally, we bind our transcription approach with a novel source-separation approach, based on explicit source decomposition with disjoint decoders. The idea behind our method is to use the knowledge previously acquired on individual instruments in order to ease their recognition in the mixture signal. A novel form of inference network is trained on the product space of the decoders latent space, with additional latent dimensions that performs Bayesian inference directly over mixture coefficients.

\section{State of the art}\label{sec:stateoftheart}
Here, we provide a brief state-of-the-art of the most common approaches for musical transcription. Then, we introduce \textit{variational auto-encoders} and detail their use for cross-modal inference and generation.

\subsection{Automatic music transcription}

Automatic music transcription (AMT) aims at closing the gap 
between acoustic music signals and their corresponding musical notation. The main problem in AMT is detecting multiple and possibly overlapping in time pitches. Classical approaches for pitch and multi-pitch extraction are mostly based on spectral or spectral analysis using fundamental harmonics localization \cite{drugman2011joint}, such as the Yin algorihtm \cite{de2002yin}. As these methods were originally conceived for monophonic signals, their extension to multi-pitch estimation contexts often implies recursive processes (multi-fundamental recognition, harmonic subtraction) that reduce their efficiency. In parallel, other methods relying on spectrogram factorization have been proposed. These are based on the decomposition of the spectrogram into a linear combination of non-negative factors. These include Non-negative Matrix Factorisation (NMF) \cite{Lee1999} or probabilistic latent component analysis (PLCA) \cite{Shashanka2008}. However, spectrogram factorization methods usually fail to identify a global optima, a limitation which led many researchers to hypothesize the need for supplementary external knowledge to attain more accurate decompositions \cite{5957256,4959583}.

Recently, deep learning approaches have been proposed to address the multi-pitch detection problem. For instance, piano transcription task has been tackled via a variety of neural networks in 
\cite{journals/corr/KelzDKBAW16,Sigtia:2016:ENN:2992480.2992488, hawthorne2017onsets,hawthorne2018enabling}. Interestingly, the MusicNet dataset \cite{thickstun2017learning} includes multi-instrument music conveniently structured to address polyphonic music transcription. Finally, a method based on convolutional neural networks is presented in \cite{Bittner2017DeepSR}, which aims at learning meaningful representations allowing accurate pitch approximation in polyphonic audio recordings.

\subsection{Generative models and variational auto-encoders}
\subsubsection{Variational inference}
\textit{Generative models} define a class of unsupervised machine learning approaches aiming to recover the probability density $p(\mathbf{x})$ underlying a given dataset. This density is usually conditioned on another set of random variables $\mathbf{z}$, called \textit{latent variables}. This set acts as a higher-level representation that controls the generation in the data domain. Formally, generative models can be described as modeling the joint probability $p(\mathbf{z,x}) = p(\mathbf{x|z})p(\mathbf{z})$, where $p(\mathbf{z})$ acts as a Bayesian \textit{prior} over the latent variables. The \textit{generative process} takes a latent position $\mathbf{z}$ to produce the corresponding probability density $p(\mathbf{x|z})$ in the data domain. Conversely, we also want to estimate the \textit{posterior} distribution $p(\mathbf{z|x})$, that gives the latent distribution corresponding to a data sample $\mathbf{x}$. Retrieving this posterior distribution from a given generative process is called \textit{Bayesian inference}, and is known to be a very robust inference framework. Unfortunately, this inference is generally intractable for complex distributions or requires limiting assumptions on both generative and inference processes. \textit{Variational inference} (VI) is a framework that overcomes this intractability by turning Bayesian inference to an optimization problem \cite{jaakkola2000bayesian}. To do so, variational inference posits a parametric distribution $q(\mathbf{z|x})$ that can be freely designed, and optimizes this distribution to approximate the real posterior $p(\mathbf{z|x})$. This optimization is performed thanks to the following bound 
\begin{equation}
\label{elbo}
\log{p(\mathbf{x})} \geq \mathbb{E}_{q(\mathbf{z|x})}\big[p(\mathbf{x|z})\big] + D_{KL} \big[ q(\mathbf{z|x}) \Vert p(\mathbf{z}) \big] = \mathcal{L}_{\mathrm{ELBO}}(q)
\end{equation}
where $D_{KL}$ denotes the Kullback-Leibler divergence. We can see that maximizing the right term of this inequality inherently optimizes the evidence $p(\mathbf{x})$ of our model. This bound, called the \textit{Evidence Lower-BOund} (ELBO), can be interpreted as the sum of a likelihood term $p(\mathbf{x|z})$ and of a divergence term that enforces the approximated posterior $q(\mathbf{z|x})$ to match the prior $p(\mathbf{z})$. This variational formulation is less restrictive than direct Bayesian inference, as it only requires the tractability of these two terms. Thus, we are able to model complex dependencies between $\mathbf{x}$ and $\mathbf{z}$ for both $p_\theta(\mathbf{x|z})$ and $q_\phi(\mathbf{z|x})$ while retaining the benefits of a Bayesian formulation \cite{bishop}.

\subsubsection{Variational auto-encoder and cross-modal learning}
To define the approximate distribution, we can model both generative and inference models as normal distributions 
\begin{align*}
    q(\mathbf{z|x}) & = \mathcal{N}\big(\boldsymbol{\mu}_q(\mathbf{x}), \boldsymbol{\sigma}_q^2(\mathbf{x}) \big)\\
    p(\mathbf{x|z}) & = \mathcal{N}\big(\boldsymbol{\mu}_p(\mathbf{z}), \boldsymbol{\sigma}^2_p(\mathbf{z}) \big)
\end{align*}
such that parameters $(\boldsymbol{\mu}_q, \boldsymbol{\sigma}_q^2)$ and $(\boldsymbol{\mu}_p, \boldsymbol{\sigma}_p^2)$ are respectively obtained by deterministic functions $f_\theta(\mathbf{x}; \theta)$ and $g_\phi(\mathbf{z}; \phi)$. When these functions are parametrized as neural networks, we obtain the original \textit{Variatonal Auto-Encoder} (VAE) formulation proposed by Kingma \& al. \cite{kingma2013auto}. The prior is usually defined as an isotropic normal distribution 
$p(\mathbf{z}) = \mathcal{N}(\mathbf{0, I})$, which acts as a regularizer to enforce the independence of latent dimensions. Similar to auto-encoding architectures, 
$f_\theta(\mathbf{x}; \theta)$ and $g_\phi(\mathbf{z}; \phi)$ are respectively called the \textit{encoder} and the \textit{decoder} of the system. These functions are jointly trained until convergence on parameters $\{\theta, \phi \}$ with a back-propagation algorithm. Despite the apparent simplicity of its formulation, this system allows very expressive encoding and generative processes while providing a highly structured latent space, whose smoothness is provided by the $D_{KL}$ reconstruction term. \\


\section{Cross-modal VAE for music transcription}\label{sec:bimodal}

\subsection{Signal/symbol transfer through shared latent spaces }
In this paper, we propose to reformulate the audio transcription problem as the estimation of a joint probability density $p(\mathbf{x,y})$, where $\mathbf{x}$ represents the spectral information of the analyzed audio signal and $\mathbf{y}$ represents the corresponding set of symbolic information. Previous works showed the efficiency of VAEs for audio processing when used on spectral frames, in terms of both representational and generative abilities \cite{esling2018generative, bitton2018modulated}. However, we intend here to estimate not only the probability density $p(\mathbf{x})$, but also the joint probability density $p(\mathbf{x,y})$. Considering $\mathbf{y}$ as label information, some approaches proposed to include an additional discriminator on the latent space, that is jointly trained during the learning process \cite{kingma2014semi}. Here, we take inspiration from the SCAN approach proposed by Higgins \& al., that trains a mirrored VAE on symbolic data whose latent representation is constrained to match the latent space obtained from the signal VAE \cite{higgins2017scan}. Hence, modelling our symbolic information as binary vectors $\mathbf{y} = [y_1, ..., y_L]$, we can train this VAE over the label space 
\begin{align*}
    q(\mathbf{z|y})  &= \mathcal{N}\big(\boldsymbol{\mu}_q(\mathbf{y}), \boldsymbol{\sigma}_q^2(\mathbf{y}) \big)\\
    p(\mathbf{y|z})  &= \prod_{i=1}^L \mathcal{B}\big(\mu_{p,i}(\mathbf{z})) \ \ \text{or} \ \ \prod_{i=1}^L \text{Cat}(\boldsymbol{\mu}_{p,i}(\mathbf{z}))
\end{align*}
where $\mathcal{B}\big(\mu_{p,i}(\mathbf{z}))$ denotes a Bernouilli distribution of mean $\mu_{p,i}$ for binary symbols, and $\text{Cat}(\boldsymbol{\mu}_{p,i}(\mathbf{z}))$ denotes a Categorical distribution of classwise probabilities $\boldsymbol{\mu}_{p,i}$ in the case of multi-label symbols. We enforce its latent representation to fit the one obtained with the signal VAE by adding a term to the ELBO
\begin{equation}
\label{scan_elbo}
\mathcal{L}_{\text{scan}}(q) =  \mathcal{L}(q) + D_{KL} \big[ q(\mathbf{z|x}) \Vert q(\mathbf{z|y}) \big] 
\end{equation}
such that the latent distributions provided by the two inference processes match for a given pair $(\mathbf{x,y})$. The ordering of terms in the Kullback-Leibler divergence is chosen such that the distribution $q(\mathbf{z|y})$ is forced to cover the whole mass of $q(\mathbf{z|x})$. Hence, the correct label for a given $\mathbf{x}$ is encouragqed even for low-probability areas of $q(\mathbf{z|x})$. Both VAEs are jointly trained, so that the latent representation obtained is a compromise between both auto-encoder performances. It should be noted that, as both signal and symbolic VAEs are independent, we are still able to perform semi-supervised learning for incomplete pairs $(\mathbf{x,y})$ by training only one of the two auto-encoders.\\

\subsection{Bidirectional signal-to-symbol mappings}
\begin{figure}
    \centering
    \includegraphics[scale=0.4]{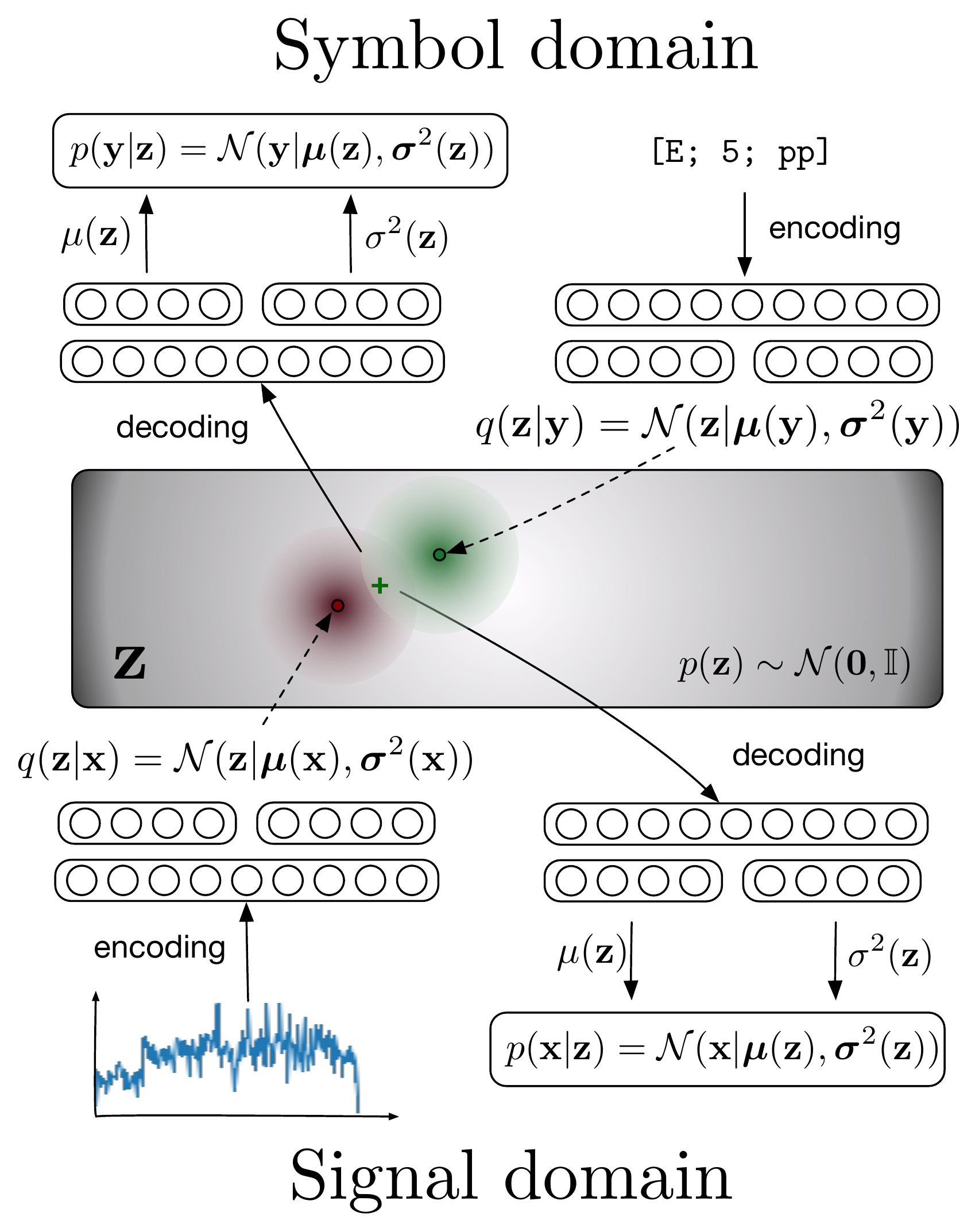}
    \caption{Multi-modal variational auto-encoding process. Our model is based on two separate variational auto-encoder, one in the signal domain and one in the symbolic domain, sharing a common latent space.}
    \label{fig:scan}
\end{figure}
Our approach extends the multi-pitch detection problem on several aspects. First, our model is inherently bi-directional as we can recover symbolic inference with the process
\[ p(\mathbf{y|x}) = p(\mathbf{y|z})q(\mathbf{z|x}) \]
This can be understood as a Bayesian formulation of audio semantic labeling. Hence, multi-pitch transcription is simply a special case of our formulation, where $\mathbf{y}$ is defined as being solely the pitch information. Furthermore, we can also naturally handle \textit{signal generation from symbolic constraints}, by taking the reverse process
\[ p(\mathbf{x|y}) = p(\mathbf{x|z})q(\mathbf{z|y}) \]
such that we can recover the appropriate spectral distribution from the symbolic data, as depicted in Fig.~\ref{fig:scan}. Another interesting property of our method is its applicability to arbitrary symbols. In this paper, we model symbolic information $\mathbf{y}$ as a triplet \texttt{[pitch class, octave, dynamics]}, where we add dynamics estimation to the pitch estimation task. Thus, we have $p(\mathbf{y|x}) = p(\mathbf{y}^p|\mathbf{z})p(\mathbf{y}^o|\mathbf{z})p(\mathbf{y}^d|\mathbf{z})$, where  each $p(\mathbf{y}^\cdot|\mathbf{z})$ is defined as a categorical distribution. We use this property to extend this method to multi-pitch applications, where $\mathbf{x}$ is a mixture signal with $M$ different sources. Hence, we formulate the symbolic information as a product $p(\mathbf{y|z}) = p(\mathbf{y_1|z})...p(\mathbf{y_M|z})$, where each $p(\mathbf{y_\cdot|\mathbf{z}})$ follows the previous specification. In addition to performing multi-pitch estimation, it also specifies the corresponding instrument if a given symbolic ordering is held during training. Finally, our formulation can be extended to polyphonic instruments in a straightforward manner. In this case, we simply replace the above conditioning by $p(\mathbf{y|z}) = p(\mathbf{y}^p|x)p(\mathbf{y}^o|\mathbf{z})$, where we define  $p(\mathbf{y}^p|\mathbf{z})$ to be a Bernoulli distribution over a one-hot pitch vector.

\section{Experiments}\label{sec:experiments}
\subsection{Datasets}
To evaluate our approach we use the Studio One Line (SOL) \cite{ballet1999studio}, a database that contains solo instrument recordings for every note across their tessitura. Each note is recording over a range of different dynamics (\textit{ff,mf,pp}). Here, we selected five instruments: violin, alto-sax, flute, C-trumpet and piano, for a total amount of 800 files. First, audio files are all resampled to a sample rate of 22050Hz. Then, we transform the raw audio data to the spectral domain by using a \textit{Non-Stationary Gabor Transform} (NSGT) \cite{velasco2011constructing}. Interestingly, this multi-resolution spectral transform allows to define custom frequency scales, while remaining invertible. Here, we use a constant-Q scale with 48 bins per octave. For each model training, we split our dataset with 80\% as training and 20\% as test sets. As our dataset is composed of monophonic signals, we randomly create instrument signal mixtures during training such that every combination is seen during the training. \\

\subsection{Models}
To show the efficiency of our proposal, we rely on VAEs with very simple architectures. Nevertheless, depending on the complexity of the input data, we adjust the dimensionality of both the latent space and hidden layers. For single-instrument models we use 32 dimensions for the latent space, and define both encoding and decoding functions for the signal VAE as 2-layers multi-layer perceptrons (MLP) with 2000 hidden units. For the symbolic auto-encoder, encoding and decoding MLPs have 2 layers and 800 hidden units. For mixtures of two different instruments, the number of hidden units for the signal encoders/decoders are set to 5000. For the mixture of three instruments, hidden layers have 5000 and 1500 units for the signal and the symbolic encoders / decoders respectively. All models are trained using the ADAM optimizer, and we use the warm-up procedure that slowly brings the regularization from 0 to 1 during the first 100 epochs. As recommended by Higgins \& al., the additional term presented in (\ref{scan_elbo}) is scaled up to a factor 10. The learning rate is first set to 1e-3, and is increasingly reduced as the derivative of the error decreases. \\

\subsection{Evaluation}
In addition to performing a standard evaluation on the test set, we also evaluate our model on a separate dataset containing recordings of flute arpeggios, scales and melodies \cite{elena_agullo_cantos_2018_1408985} with source audio files and aligned MIDI files. Unfortunately, this dataset does not provide information about symbolic dynamics, so we do not evaluate the dynamics inference on this set. We compare the efficiency of our model with results obtained from a baseline approach. To this end, we rely on an architecture similar to our model, but designed in a supervised way to emphasize the gain provided by our model. This baseline classifier first performs a Principal Component Analysis (PCA) from the signal data to perform dimensionality reduction, mocking the compression between the input data and the latent space. Then, we use a 2-layer MLP with the same amount of hidden units than the corresponding symbolical decoder, to output the desired labels. The whole system is trained on a standard cross-entropy loss. The classifier is trained until convergence with the same optimization strategy.

\section{Results}\label{sec:results}
In this section, we present the results of our methods. The source code, audio examples and additional figures and results are available on our support page \texttt{\url{https://domkirke.github.io/latent-transcription/}}.

\subsection{Signal reconstruction and transfer performances}

\begin{table}[!t]
\caption{Signal reconstruction and transfer performances}
\label{table:signal}
\centering
\centering
\scalebox{0.85}{
\begin{tabular}{c||c|c||c|c}
\hline
 & $-\log{p(\mathbf{x|z})}$ & ISD & $-\log{p(\mathbf{x|y})}$ & \textit{ISD} \\
 
\hline
Alto-Sax (Sax) &
 -694.1 & 0.093 & -416.6 & 0.177  \\
\hline

Violin (Vn) & -671.4 & 0.104 & -551.1 & 0.151 \\

\hline
Trumpet-C (TpC) & -706.9 & 0.073 &  276.71 & 0.35  \\

\hline
Flute (Fl) & -706.2 & 0.076 &  -379.2 & 0.147 \\

\hline
Piano (Pn) & -813.5 & 0.044 & -361.13 & 0.112  \\
\hline

Sax + Vn & -358.71 & 0.364 & -27.37 & 0.852  \\
\hline

Sax + Vn + Fl & 
 -268.7 & 0.624 & 692.4 & 3.813   \\
\hline
\end{tabular}
}
\end{table} 

\begin{figure*}[ht]
    \centering
    \label{figure:reconstruction}
    \includegraphics[scale=0.55]{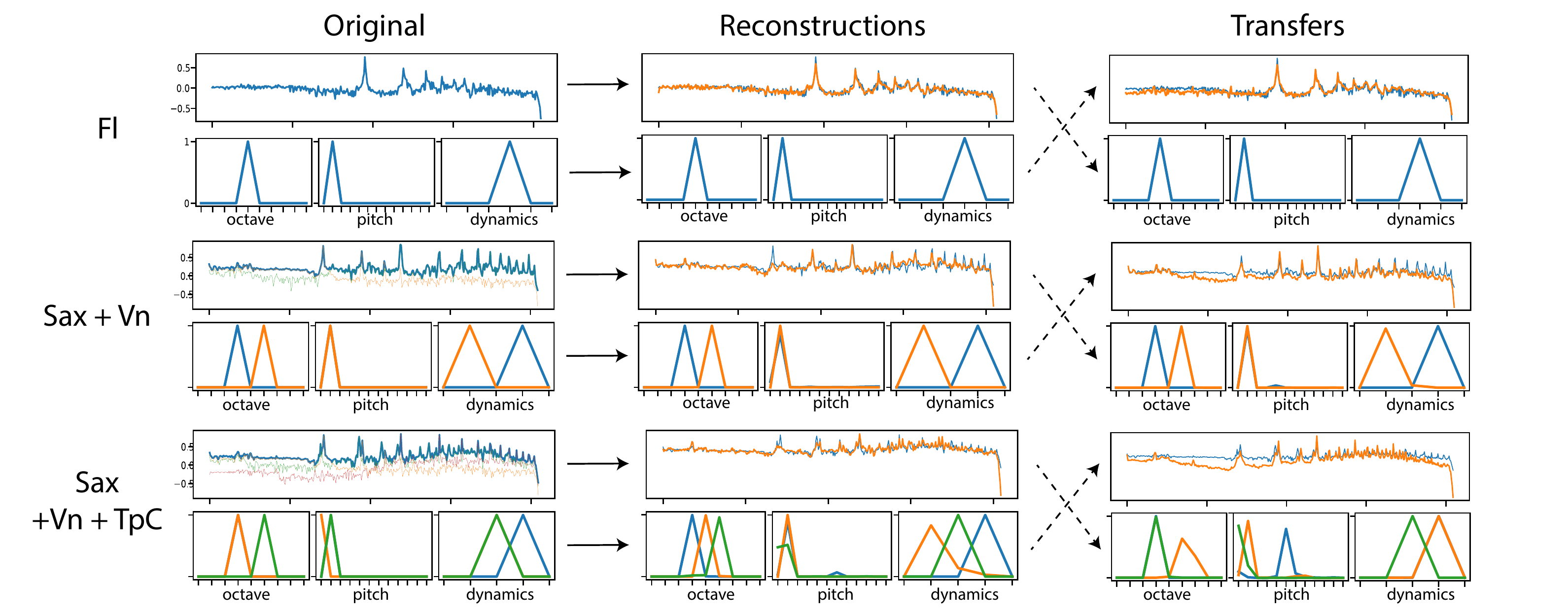}
    \caption{Signal \& symbolic reconstruction samples in 1-instrument, 2-instruments and 3-instruments settings. The first column shows the original spectral \& symbolic contents : for spectra, the blue line represents the final mixture, and the thinner lines the different components of the mixture. For symbols, labels are grouped by family and are symbolized by a peak at the correct label. The second column represents reconstruction results. For spectra, the orange line represents the reconstructed spectra (the original spectra in blue is left for comparison). Regarding the symbolic reconstructions, the corresponding categorical distributions are displayed right to the original one-hot vectors. The third column finally shows the transfer results, where we decode a latent position given by the encoder of the other domain. }
\end{figure*}

First, we analyze the results obtained on the SOL examples. Signal reconstruction and transfer scores are provided in Table~\ref{table:signal}, relying on two evaluation metrics. The first metric is the log-likelihood of the original spectrum with respect to the distribution decoded by the model. The second is the Itakura-Saito Divergence (ISD), a metric that reflects the perceptual dissimilarity between the original and reconstructed spectrum \cite{6797100}. Both scores are presented for signal-to-signal reconstruction (left) and symbol-to-signal inference (right). In addition to these scores, reconstruction examples are depicted in Fig.~\hyperref[figure:reconstruction]{2}. We can see that performances in both signal reconstruction and transfer decrease with the number of instruments, as the complexity of the incoming signal increases. Both reconstruction and signal-to-transfer scores are almost perfect in the case of solo instruments, providing convincing and high-quality sound samples generation. In the case of mixtures of two or more instruments, reconstruction scores maintain an acceptable performance, but symbol-to-signal transfer scores clearly decrease. This observation correlates with the decrease of performance observed in the symbolic domain, as discussed in the following sub-section.

\subsection{Symbolic inference performances}

\begin{table}[!t]
\caption{Symbolic inference reconstruction and classification results (successively pitch, octave and dynamics). Scores without parenthesis are reconstruction scores obtained within the symbolic domain, while scores in parenthesis are obtained when performing transfer from the signal domain}
\label{table:symbol}
\centering
\scalebox{0.60}{
\begin{tabular}{||c||c|c|c|c||c}
\hline
&  & $-\log{p(\mathbf{y|z})}$  & Success Ratio (\%) & \textit{loose} (\%) & Baseline (\textit{loose}) \\
 
\hline
Sax & \begin{tabular}{@{}c@{}} p \\ o \\ d \end{tabular} & 
\begin{tabular}{@{}c@{}} -1.0 \textit{(-1.0)} \\ -1.0 \textit{(-1.0)}  \\ -1.0 \textit{(-1.0)} \end{tabular} &
\begin{tabular}{@{}c@{}} 100\% \textit{(100\%)} \\ 100\% \textit{(100\%)}  \\ 100\% \textit{(100\%)} \end{tabular} &
\begin{tabular}{@{}c@{}} - \\ -  \\ - \end{tabular} & \begin{tabular}{@{}c@{}}  \textit{94\%} \\ \textit{97\%} \\ \textit{46\%} \end{tabular}  \\

\hline
Vn & \begin{tabular}{@{}c@{}} p \\ o \\ d \end{tabular} & 
\begin{tabular}{@{}c@{}} -1.0 \textit{(-1.0)} \\ -1.0 \textit{(-1.0)}  \\ -1.0 \textit{(-1.0)} \end{tabular} &
\begin{tabular}{@{}c@{}} 100\% \textit{(100\%)} \\ 100\% \textit{(100\%)} \\ 100\% \textit{(100\%)} \end{tabular} &
\begin{tabular}{@{}c@{}} - \\ -  \\ - \end{tabular} &
\begin{tabular}{@{}c@{}}\textit{ 89.8\%} \\ \textit{99.0\%} \\ \textit{35.3\%} \end{tabular}  \\

\hline
TpC & \begin{tabular}{@{}c@{}} p \\ o \\ d \end{tabular} & 
\begin{tabular}{@{}c@{}} -1.0 \textit{(-1.0)} \\ -1.0 \textit{(-1.0)}  \\ -0.998 \textit{(-1.0)} \end{tabular} &
\begin{tabular}{@{}c@{}} 99.9\% \textit{(100\%)} \\ 100\% \textit{(100\%)}  \\ 99.7\% \textit{(100\%)} \end{tabular} &
\begin{tabular}{@{}c@{}} - \\ -  \\ - \end{tabular} &
\begin{tabular}{@{}c@{}} \textit{76.1\%} \\ \textit{99.8\% }\\ 47.8\%\textit{} \end{tabular}  \\

\hline
Fl & \begin{tabular}{@{}c@{}} p \\ o \\ d \end{tabular} & 
\begin{tabular}{@{}c@{}} -1.0 \textit{(-1.0)} \\ -1.0 \textit{(-1.0)}  \\ -1.0 \textit{(-1.0)} \end{tabular} &
\begin{tabular}{@{}c@{}} 100\% \textit{(100\%)} \\ 100\% \textit{(100\%)}  \\ 100\% \textit{(100\%)} \end{tabular} &
\begin{tabular}{@{}c@{}} - \\ -  \\ - \end{tabular}&
\begin{tabular}{@{}c@{}} \textit{52.4\%} \\ \textit{81.8\%} \\ \textit{41.4\%} \end{tabular}  \\

\hline
Pn & \begin{tabular}{@{}c@{}} p \\ o \\ d \end{tabular} & 
\begin{tabular}{@{}c@{}} -1.0 \textit{(-1.0)} \\ -1.0 \textit{(-1.0)}  \\ -1.0 \textit{(-0.999)} \end{tabular} &
\begin{tabular}{@{}c@{}} 100\% \textit{(100\%)} \\ 100\% \textit{(100\%)}  \\ 99.9\% \textit{(100.0\%)} \end{tabular} &
\begin{tabular}{@{}c@{}} - \\ -  \\ - \end{tabular} &
\begin{tabular}{@{}c@{}} 51.6\% \\ 63.9\% \\ 40.0\% \end{tabular}  \\

\hline
Sax + Vn & \begin{tabular}{@{}c@{}} p \\ o \\ d \end{tabular} & 
\begin{tabular}{@{}c@{}} -0.534 \textit{(-0.871)} \\ -0.782 \textit{(-0.980)}  \\ -0.712 \textit{(-0.939)} \end{tabular} &
\begin{tabular}{@{}c@{}} 54.0\% \textit{(87.9\%)} \\ 84.6\% \textit{(99.2\%)}  \\ 74.4\% \textit{(95.9\%)} \end{tabular} &
\begin{tabular}{@{}c@{}} 62.6\% \textit{(81.6\%)} \\ 94.9\% \textit{(88.7\%)} \\ 82.4\% \textit{(66.3\%)} \end{tabular} &
\begin{tabular}{@{}c@{}} \textit{65.3\% }  \\ \textit{79.1\%}  \\\textit{ 52.0\% } \end{tabular}  \\

\hline
Sax + Vn + TpC & \begin{tabular}{@{}c@{}} p \\ o \\ d \end{tabular} & 
\begin{tabular}{@{}c@{}} -0.381 \textit{(-0.725)} \\ -0.377 \textit{(-0.641)}  \\  -0.347 \textit{(-0.616)} \end{tabular} &
\begin{tabular}{@{}c@{}} 38.6\% \textit{(75.0\%)} \\ 42.4\% \textit{(67.8\%)}  \\ 34.6\% \textit{(62.4\%)} \end{tabular} &
\begin{tabular}{@{}c@{}} 62.6\% \textit{(84.5\%)} \\ 79.3\% \textit{(88.7\%)} \\ 66.9\% \textit{(69.5\%)} \end{tabular} &
\begin{tabular}{@{}c@{}} \textit{56.6\%} \\ \textit{62.3\%}  \\ \textit{41.2\% } \end{tabular}  \\
\hline
\end{tabular}
}
\end{table}

Here, we evaluate the performances of our model in the symbolic domain. We provide in Table~\ref{table:symbol} four different classification scores, separately for each family of labels: octave, pitch class and dynamics. In the case of multi-instrument mixtures, these losses are averaged over every instrument of the mixture. Every column (except for the baseline) show two scores : the first are the scores obtained symbol-to symbol (reconstruction), and the second within parenthesis are the ones obtained signal-to-symbol (transfer). \\

The first loss, written $-\log{p(\mathbf{y|z})}$, denotes the \textit{likelihood} of the true labels with respect to the distributions decoded by the symbolic part of the VAE. The percent scores located at the right of the likelihood correspond to classification scores, obtained by taking the highest probability of the categorical distribution and obtaining the corresponding ratio of well-classified symbols. The first column, called \textit{success ratio}, denotes the classification score obtained by the symbolical VAE. The second column, called \textit{loose ratio}, is specific in the case of mixed instruments, considering a label to be correct regardless of the instrument (we will come back to the motivation behind this score). Finally, the last column display the scores obtained by our baseline classifier, that does not have symbol-to-symbol scores.\\

We note that symbolic reconstruction and signal-to-symbol scores are almost perfect in the case of single-source signals, outperforming the equivalent baseline system. We argue that is due to two main aspects of the proposed approach. First, thanks to the reconstruction task, the construction of the latent space is organized to reflect the inner structure of both signal and symbol domains. The latent space can be thus understood as a feature space, carrying higher-level information that allow signal/symbolic coupling to be more efficient. Second, the Bayesian approach matching the latent spaces allows a smoother and more efficient mapping than a deterministic approach, that would just provide pairwise mappings between incoming examples. \\ 

In the case of instrument mixtures, scores are decreasing as the complexity of both the spectrum and symbolic distributions increase. While the system still performs convincingly with two instruments mixtures, it struggles with mixtures of more than three instruments. We argue that this is partly due to a combinatorial problem, as can be seen when analyzing the \textit{loose classification ratio} : indeed, it becomes harder for the model to accurately affect a label to the corresponding instrument as the number of different possible sources increase. This can be seen with the \textit{loose ratio} in table \ref{table:symbol} : where the classification ratio significantly increase if the label is considered correct regardless the affected instrument. This effect can also be seen figure 2, where some peaks are correct but unfortunately distributed to the wrong instrument. A more subtle strategy to tackle this effect has to be considered ; we leave this to a future work. \\

\subsection{Monophonic flute transcription}

\begin{table}[!t]
\caption{Results obtained on an external flute dataset}
\label{table:flute}
\centering
\begin{tabular}{|c|c||c|c|}
\hline
 Likelihood & ISD & Class. Ratio & Baseline  \\
 \hline
 2648 (\textit{1057}) & 1.065 (\textit{0.632}) &
 \begin{tabular}{@{}c@{}} 65.4\% \\ 81.9\% \end{tabular} & 
 \begin{tabular}{@{}c@{}} 63.8\% \\ 76.8\% \end{tabular} \\
\hline
\end{tabular}
\end{table}

Here, we analyze the results obtained with an external dataset of flute recordings, as depicted in Table \ref{table:flute}. Performances in symbolic inference is still convincing, showing that our model does not suffer from strong over-fitting. Compared to the results obtained with the reference dataset \ref{table:signal}, the reconstruction results obtained here have decreased. This is due to several points : first, we have trained the model solely on the stationary part of each instrument signals, such that the attack and release of the signal are not understood by our model. This anomaly is clearly perceptible when listening to the reconstructions. Second, a more subtle comparison between the reference dataset and this dataset showed important differences in terms of harmonic content. Specifically, a 1-octave lower harmonic is globally present in this dataset, and not in the reference one. This may explain the important decrease in octave classification, and may indicate that an increased amount of various instruments of the same type may be required to enforce the generalization of the model. \\

\section{Discussion and future works}

\subsection{Performance aspects}
We think that the efficiency of the proposed approach mainly relies on the hybridization of its learning process, that combines both unsupervised and supervised learning. Indeed, while each encoder learns to extract domain-dependant features in an unsupervised manner, latent spaces are matched by enforcing a supervised coupling of signal/symbol pairs. This process thus intends to learn transferable features, that are then used by each decoder to project them back into their respective data domains. Furthermore, this process allows the model to train on incomplete data, such that each domain's encoding / decoding functions can still be trained individually even if some signal / symbol couplings are missing. This means that the training method is scalable to bigger datasets where some symbolic information may be absent, such that incomplete data can yet be used to reinforce the reconstruction abilities of the system. \\

However, in spite of the strengths of the proposed approach, the actual state of the model suffers some issues, that we aim to tackle in the future. The first main issue is that, while the system performs well in the single-instrument case, its performance weakens with two instruments and clearly fails when applied on more. We think that this falls to several reasons. First, we think this is due to the capacity of the model, as we still use very simple systems even for complex signals like the 3-instruments case. Secondly, the complexity of the problem is such that, as we showed when comparing the loose and non-loose version of the classification ratio, the system struggles to correctly allocate the good label to the good instrument. We think that the incoming signal representation may be not precise enough to alleviate some ambiguities, as for examples in the case of octaves or fifths where instrument identification may be hard to disentangle. Furthermore, the model does not prevent instrument-wise symbolic outputs to focus on the same spectral components, and thus to perform redundant symbol predictions, and thus may also lead to a permutation problem.   \\
Finally, another issue with the proposed model is that the temporal evolution is not considered by the system. Including temporal features could bring decisive enhancements : in addition to allow full-sound generation and increased pitch and dynamics inference, it may even be mandatory for applying our model to custom symbolic dictionaries (playing modes, temporal symbols such as trills...) and provide a substantial advantage over more casual pitch-detection methods.\\

\subsection{Creative aspects}

\begin{figure}[ht]
    \centering
    \label{fig:transcription}
    \includegraphics[scale=0.35]{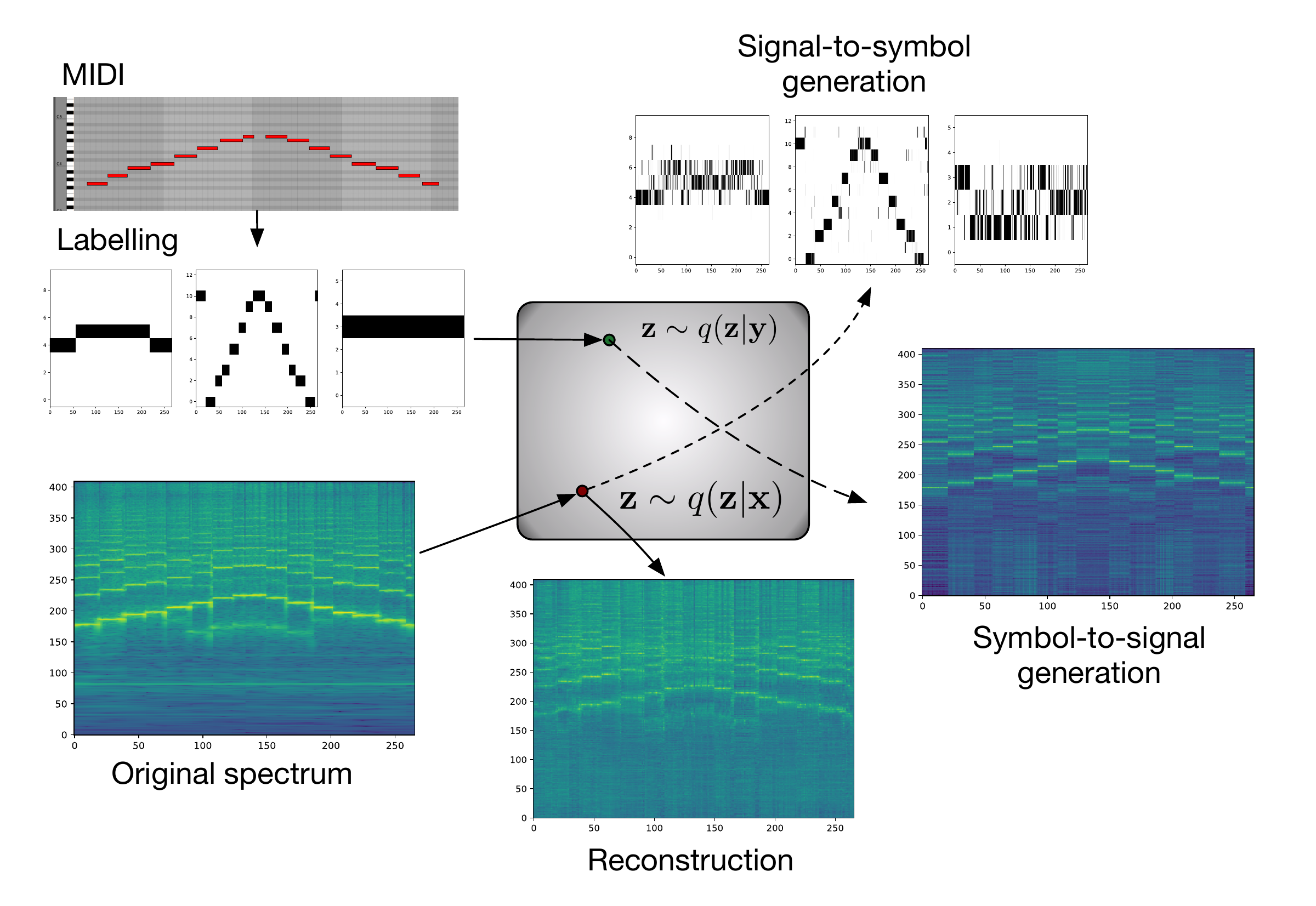}
    \caption{Diagram showing in-domain and cross-domain inference. If we give a spectrum to the signal encoder and draw from the corresponding distribution a  latent position $\mathbf{z}$, we can decode it with the signal generation network (in-domain) and obtain the reconstruction. Alternatively, we can decode it with the symbol generation network (cross-domain) to perform \textit{signal-to-symbol} generation, that here is equivalent to transcription. Reversely, if we draw a latent position from the symbol inference network with a given set of labels, we can decode it with the signal decoder to perform \textit{symbol-to-signal} generation, that allows us to "play" a MIDI file with our model.}
\end{figure}

Finally, an important aspect of the proposed model is the diversity of creative applications it provides (see figure \hyperref[fig:transcription]{3}). As generation of both symbolic and signal content is both based on the latent space, one may use it as a continuous control space and meaningfully explore it in either an unsupervised or semi-supervised fashion. Indeed, this space can be explored in a fully unsupervised manner by direct interaction: both signal and symbol information are then generated, such that the user can have a direct symbolic feedback on the data he is generating. Alternately, it can also be used in a semi-supervised fashion, constraining the navigation to the distribution inferred by a given symbol or a given sound. For example, in our case, we can directly generate a note with given pitch, octave and dynamics by inferring a distribution with the symbolic encoder, and then navigate inside it to access the diversity of signals retained under the corresponding label information. This allows us, translating first MIDI information in pitch/octave/dynamics pairs, and then transferring this symbolic information in the signal domain, to generate audio content from a MIDI file. We list below various use cases that can be carried out by our model: 

\begin{itemize}
    \item \textbf{sequence generation}: we can use a sequence of labels to recover the corresponding distribution in the latent space that we can freely sample and/or navigate, 
    \item \textbf{spectral morphing}: we can take two latent target distributions, and draw a trajectory that we can sample regularly to obtain a smooth transformation between the two target sounds,
    \item \textbf{free trajectory}: take a totally free trajectory in the latent space,
    \item \textbf{symbol extraction}: we can infer symbolic information from an incoming signal, and still train the corresponding signal encoder/decoder with the incoming data. This could be a particularly interesting feature especially in real-time contexts.
\end{itemize}

Corresponding examples for each of the above navigation strategies are given at { \tt \href{https://domkirke.github.io/latent-transcription/}{support webpage}}. Finally, also note that, in our example, the vocabulary is easy to learn, such that retrieving the underlying distribution $p(\mathbf{y})$ of the symbolic data itself is not really useful. Indeed, the different labels are all independent, and are approximately equally distributed in their own domain. However, our system can also learn on much more complex vocabularies where learning the underlying distribution in the symbolic domain itself has an interest, and thus open additional perspectives for its use in creative and/or MIR applications.

\section{Conclusion}
In this paper, we proposed a novel formulation for bijective signal/symbol translation, based on the latent space matching of domain-wise variational architectures. We studied the benefits and drawbacks of the proposed system, and concluded that while improvable this model performed well and proposed a very interesting alternative to signal-symbol algorithms, and furthermore provided additional applications that were not possible in previous models. Indeed, our method is bi-directional, and performs well in both audio-to-symbol and symbol-to-audio prediction. Furthermore, our method is compatible with any kind or arbitrary symbolic information, and is then opened to user-defined vocabularies. Besides, as our model is based on a latent space that can be considered as a continuous control space, it is also opened to diverse creative uses as sequence generation, sound interpolation, or free navigation, whether in an supervised manner or semi-supervised manner, guided with symbolic information. For future work, we plan to solve the symbolic ambiguities that raise in the case of numerous instruments, to incorporate temporal features to allow dynamical features extraction, and to design user interfaces to make our model compatible with artistic practises. 

\nocite{*}
\bibliographystyle{IEEEbib}
\bibliography{biblio.bib} 
\balance

\end{document}